\documentclass[11pt]{article}

\usepackage[final]{acl}

\usepackage{times}
\usepackage{latexsym}

\usepackage{graphicx}
\usepackage{multirow}
\usepackage{amssymb} % for \checkmark
\usepackage{makecell}
\usepackage{tabularx}
\usepackage{placeins}


\newcolumntype{Y}{>{\centering\arraybackslash}X}
\usepackage{array}

\usepackage[T1]{fontenc}
\usepackage[utf8]{inputenc}

\usepackage{microtype}

\usepackage{inconsolata}

\usepackage{graphicx}

\usepackage{framed}
\usepackage{time}
\usepackage[most]{tcolorbox}
\usepackage{enumitem}
\title{Contextual Observer Grounding: \\
Evaluating Situated Spatial Reasoning in Vision-Language Models}

\author{
  Mimo Shirasaka \\
  The University of Tokyo \\
  Tokyo, Japan \\
  \small\texttt{m-shirasaka@g.ecc.u-tokyo.ac.jp} \\\And
  Haochen Zhang \\
  Carnegie Mellon University \\
  Pittsburgh, PA, USA \\
  \small\texttt{haochen4@andrew.cmu.edu} \\\And
  Yonatan Bisk \\
  Carnegie Mellon University \\
  Pittsburgh, PA, USA \\
  \small\texttt{ybisk@cs.cmu.edu}}

\begin{document}
\maketitle
\begin{abstract}
Reasoning over language instructions in embodied tasks such as robotics often requires understanding spatial relations from a speaker's situated perspective. Humans infer such perspectives from shared environmental knowledge, activity context, and commonsense. Recent vision-language models (VLMs) appear capable of spatial reasoning, but their ability to infer a speaker's viewpoint from contextual cues and interpret situated spatial relations from that viewpoint remains unclear. We call this capability \textbf{contextual observer grounding}. To study this capability, we construct the Point-of-View Benchmark (\textbf{POVBench}), a dataset of 3D scenes and queries that disentangles \textit{Inferred}, \textit{Stated}, and \textit{Given} forms of observer grounding in natural embodied communication. Given multi-view observations and a natural-language sentence, models must localize unseen or underspecified targets from situated spatial and contextual cues. Across state-of-the-art VLMs, localizing targets from directional language remains challenging, even when observer grounding is made explicit. We find that explicit breakdowns of observer-relative spatial reasoning improve target localization. Our project page is available at \url{https://mimo-owl.github.io/POVBench/}.

% This document is a supplement to the general instructions for *ACL authors. It contains instructions for using the \LaTeX{} style files for ACL conferences.
% The document itself conforms to its own specifications, and is therefore an example of what your manuscript should look like.
% These instructions should be used both for papers submitted for review and for final versions of accepted papers.
\end{abstract}

\section{Introduction}

\begin{figure}[t]
  \includegraphics[width=\columnwidth]{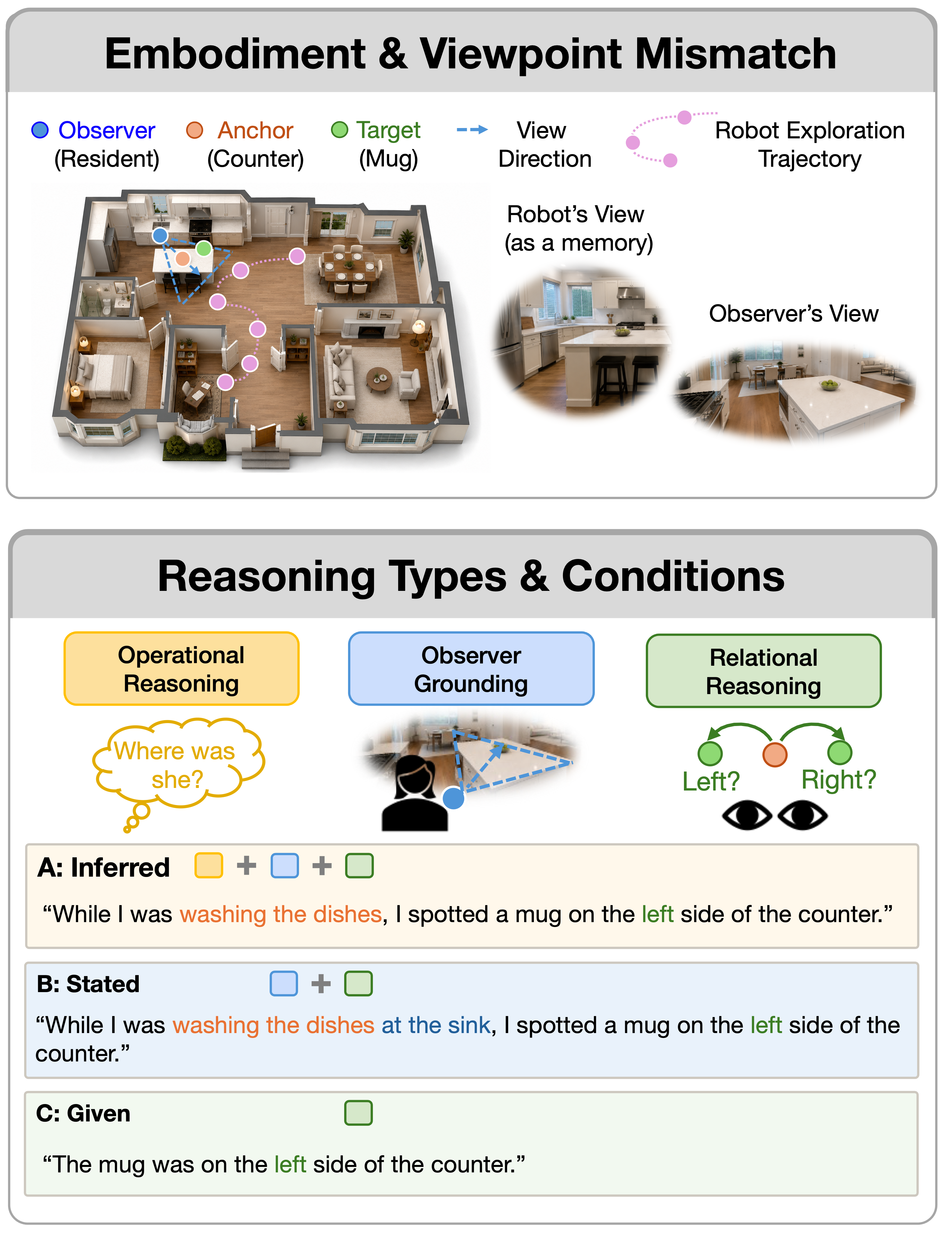}
  \caption{Overview of contextual observer grounding. Models infer the
observer's location (\textit{operational reasoning}), reconstruct their
perspective (\textit{observer grounding}), and localize the target
(\textit{relational reasoning}). Conditions \textit{A: Inferred},
\textit{B: Stated}, and \textit{C: Given} require three, two, and one of
these steps, respectively.}
  \label{fig:teaser}
\end{figure}

Humans rarely communicate through fully explicit descriptions of spaces and events. In everyday communication, people often omit information by relying on shared environmental knowledge, activity context, and assumptions that their viewpoint can be inferred. For example, having visited a friend's house, you can picture the scene simply from hearing ``When I woke up this morning, I noticed my book had fallen to the left of the lamp,'' even without sharing the speaker's exact viewpoint. In this way, humans efficiently infer situated spatial configurations by combining past visual experiences with commonsense knowledge. This ability is also fundamental for embodied AI systems that must communicate and collaborate naturally with humans, as observers often give instructions from their own viewpoint and leveraging spatial priors is important for targeted and efficient execution. For instance, if someone says ``I forgot to grab the mug I used while watching a movie yesterday,'' another person can infer that the mug is likely on the table near the sofa across from the TV. In contrast, prior work on spatial grounding generally assumes that the referring expression explicitly identifies the target through object attributes or spatial relations, with relevant perspective information typically given or directly expressed~\cite{achlioptas2020referit3d, chen2020scanrefer, zantout2025sort3d, xu2026s}.

Recent vision-language models (VLMs) have demonstrated strong performance on visual question answering and some embodied tasks~\cite{yang2025qwen3, comanici2025gemini, singh2025openai}. In particular, extensive research has explored visual perspective-taking benchmarks~\cite{achlioptas2020referit3d, ma2022sqa3d, yang2024thinkinspace, thompson2025rem}, and object-centric reasoning for navigation from exploration image sequences~\cite{min2024situated, ku2020room, anderson2018vision}. However, many existing approaches either provide the observer viewpoint explicitly, make the scene to be reasoned about directly available, or localize a target that has already been observed. Thus there is still some misalignment with the implicit contextual reasoning humans do in 3D space and it is unclear whether existing models can handle the situated contextual reasoning that humans perform with ease.

In this work, we define this problem as \textbf{contextual observer grounding}: the task of inferring from which viewpoint a speaker observed a target object using activity context and environmental cues, and then interpreting the resulting situated spatial relation (Figure~\ref{fig:teaser}). Importantly, solving such tasks should not require direct observation of the target scene itself.  Instead, the model must infer an unseen observer-relative scene pieced together from previous environment observations.

To systematically evaluate this capability, we introduce the \textbf{Point-of-View Benchmark (POVBench)}, an embodied spatial reasoning benchmark constructed using ProcTHOR-generated environments. POVBench disentangles three forms of observer grounding commonly present in natural communication: (\textit{i}) \textit{Inferred}, where the observer's viewpoint is not explicitly specified but must be inferred from activity context and environmental cues in the sentence; (\textit{ii}) \textit{Stated}, where the observer grounding is explicitly specified in the sentence; and (\textit{iii}) \textit{Given}, where the observer viewpoint is directly provided through an observer-view image. In each task, models receive first-person visual observations together with a natural-language sentence, and must infer situated spatial relations to predict the image coordinates of unseen or underspecified target objects. Keeping the query construction fixed across these conditions
turns them into a diagnostic decomposition: \textit{Inferred} versus \textit{Stated} tests the
cost of recovering observer grounding from contextual cues, whereas \textit{Stated}
versus \textit{Given} tests the cost of reconstructing an observer-relative relation without the observer view.

We conduct a comprehensive evaluation on POVBench across multiple VLMs, including proprietary and open-weight models as well as dedicated pointing architectures. Performance remains low across all three conditions, including when the observer viewpoint is explicitly \textit{Stated} or \textit{Given}. Errors differ modestly between the \textit{Inferred} and \textit{Stated}, suggesting that the challenge lies not only in identifying the observer viewpoint but also in resolving observer-relative spatial relations. Our contributions are summarized as follows:

\begin{itemize}
[leftmargin=1em,itemsep=0.15em,topsep=0.2em]

    \item We introduce \textbf{contextual observer grounding}, a form of embodied spatial reasoning that requires inferring observer-perspective spatial relations from contextual and environmental cues, and propose \textbf{POVBench}, a benchmark for evaluating this capability in embodied settings.
    
    \item We conduct a comprehensive evaluation and failure analysis across multiple VLMs, and show that current systems struggle to consistently reconstruct situated spatial relations, even when observer grounding is explicitly provided.

    \item We show that explicit step-by-step reasoning over observer-relative spatial relations within the scene improves localization performance for target objects, suggesting a potential direction for embodied spatial reasoning.

\end{itemize}

\section{Related Work}
\subsection{3D Spatial Reasoning}

Recent advances in VLMs have increasingly expanded spatial reasoning beyond static image understanding toward embodied observations, sequential exploration, and 3D scene understanding. Early benchmarks such as Visual Spatial Reasoning~\cite{liu2023vsr} and What’sUp~\cite{kamath2023whatsup} evaluate fundamental spatial relations including left/right, above/below, and near/far between objects in static images, while more recent work such as Spatial-DISE~\cite{huang2026spatialdise} studies more diverse and compositional forms of spatial reasoning. Other work extends spatial reasoning into 3D and embodied environments. SpatialVLM~\cite{chen2024spatialvlm} investigates large-scale spatial reasoning in metric 3D space, while ScanRefer~\cite{chen2020scanrefer} and ReferIt3D~\cite{achlioptas2020referit3d} study language-guided object localization in 3D scenes. In particular, the NR3D subset of ReferIt3D contains natural-language references to a target object, which sometimes specifies a perspective. OpenEQA~\cite{majumdar2024openeqa}, GraphEQA~\cite{saxena2024grapheqa}, and Explore-until-Confident~\cite{ren2024exploreuntilconfident} further examine embodied question answering over explored environments using semantic spatial memory and active exploration. 

\subsection{Perspective-Taking and Observer Grounding}

Existing benchmarks for situated reasoning address complementary aspects of contextual observer grounding. SQA3D~\cite{ma2022sqa3d} specifies a position or perspective of an agent and asks a corresponding spatial reasoning question. VSI-Bench~\cite{yang2024thinkinspace} evaluates a broad set of configurational, metric, and temporal spatial questions over egocentric video. MindCube~\cite{wang2026spatialmental} evaluates whether VLMs can construct spatial mental models from a small set of views, including under perspective changes and hypothetical movements. More recently, ViewSpatial-Bench~\cite{li2025viewspatialbench} and All-Angles Bench~\cite{yeh2025allanglesbench} evaluate multi-perspective and multi-view spatial reasoning by explicitly specifying the observer viewpoint. Related work on physical commonsense, such as PIQA~\cite{bisk2020piqa}, probes whether language models can reason about plausible physical interactions from context, a task that remains challenging for current models. These benchmarks establish the importance of situated reasoning, sequential observations, or unobserved space, but their task formulations do not enforce naturalness of language nor rely on implicit descriptions to predict both a perspective and target location for an imagined object not directly observed.

Other prior work on visual perspective-taking requires spatiotemporal reasoning or reasoning through various perspectives of the same scene. SAW-Bench~\cite{li2026sawbench} evaluates observer-relative situated awareness over a trajectory in real-world egocentric videos, while APC~\cite{lee2025perspective} studies explicit perspective transformation through intermediate scene abstractions of a scene. We propose POVBench as a coupled process of inferring an implicit observer location from contextual cues and reconstructing the observer's perspective from a series of multi-view exploration observations, without a direct view from the observer's viewpoint. We regard this as a more realistic formulation of a scenario for an embodied agent in user environments such as households.

\section{Contextual Observer Grounding}
\subsection{Task Formulation}

We formalize \textbf{contextual observer grounding} as follows.
Given a natural language sentence $S$ describing a scene observation and a set of $n$ first-person images $\mathcal{I} = \{I_0, \ldots, I_{n-1}\}$ captured during robot exploration of a home, a model must predict an image index $k \in \{0, \ldots, n-1\}$ and a pixel coordinate $(x, y)$ within $I_k$, indicating the estimated location of a target object.

This formulation poses two coupled challenges.
First, spatial language is situated: expressions such as ``to the left of the TV'' are grounded not in an absolute frame but in the observer's egocentric frame at the time of observation, so the model must first recover the observer's viewpoint before interpreting the relation. The target is therefore underspecified by the sentence alone.
Second, the target is not observed directly: because human-centric environments are dynamic, past observations need not reflect the current scene, so the target is typically absent from $\mathcal{I}$ and only the anchor object appears. The model must therefore infer where the target should be, relative to the visible anchor object and from the recovered viewpoint, rather than detect it directly.

\subsection{Taxonomy}
We identify three axes of embodied perspective reasoning, which POVBench disentangles across three conditions:

\begin{itemize}[leftmargin=1em,itemsep=0.15em,topsep=0.2em]
\item \textbf{Operational reasoning}: inferring the furniture or equipment the observer was interacting with and thus their spatial position from commonsense knowledge of human activities and environments and implicit context.
\item \textbf{Observer grounding}: determining the observer's perspective at the time of observation and reconstructing their imagined observation from multi-view images.
\item \textbf{Relational reasoning}: interpreting the spatial language in $S$ relative to the observer's viewpoint to localize the target object with respect to a named anchor object.
\end{itemize}

\subsection{Forms of Observer Grounding}
Natural language descriptions vary in how explicitly they convey the observer's viewpoint. Based on the three types of reasoning defined above, we define three conditions that systematically control this, forming a graded difficulty axis. From \textit{A: Inferred} to \textit{C: Given}, the sentence supplies progressively more explicit grounding, removing one reasoning step at each level. \textit{Inferred}, which supplies the least grounding, is the most demanding condition and our central focus.

\begin{itemize}[leftmargin=1em,itemsep=0.15em,topsep=0.2em]

\item \textbf{A: Inferred}: A set of first-person exploration images is available, but the sentence gives only an activity and/or environmental cues, without explicitly stating an object or position. The model must infer where the observer was situated from contextual cues, reconstruct the observer's perspective from that inferred location, and then resolve the relation, requiring operational, observer grounding, and relational reasoning.

\item \textbf{B: Stated}: A set of first-person exploration images is available, and the sentence explicitly names an object the observer was interacting with. The model must reconstruct the observer's perspective from this stated object and then resolve the relation, requiring observer grounding and relational reasoning.

\item \textbf{C: Given}: The observer's viewpoint is provided directly as a corresponding first-person image, and the sentence states the spatial relation between the target and a named anchor object. This requires only relational reasoning.

\end{itemize}

\section{POVBench}
\begin{figure}[t]
  \includegraphics[width=\columnwidth]{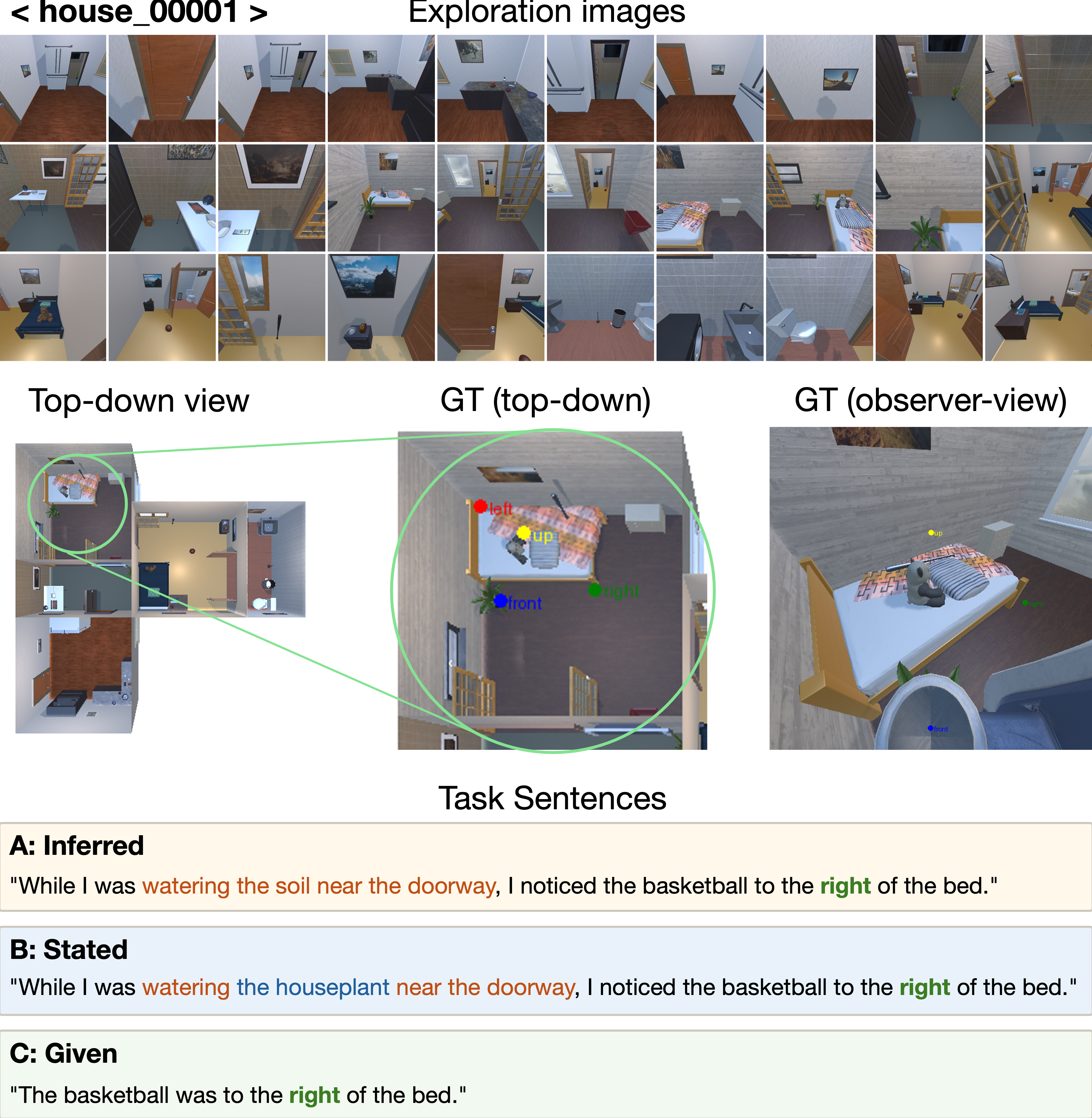}
  \caption{
    Overview of POVBench. Multi-view exploration images are collected along a simulated robot trajectory in ProcTHOR. For each observer--anchor pair, directional targets are defined around the anchor object from the observer's viewpoint. Three grounding conditions (\textit{A}, \textit{B}, \textit{C}) vary in how explicitly the sentence specifies the observer viewpoint.
    }
  \label{fig:dataset_overview}
\end{figure}

Figure~\ref{fig:dataset_overview} shows an example of the POVBench data.

\subsection{Environment Setup}
\label{sec:env}

We generate environments using ProcTHOR~\cite{deitke2022procthor}, a procedural house generation framework built on AI2-THOR~\cite{kolve2017ai2thor}. Its controllable household layouts and precise scene geometry enable scalable query generation and geometric validation of target locations. The benchmark consists of procedurally generated single-story houses with diverse layouts spanning living rooms, kitchens, bedrooms, and bathrooms. Each environment is populated with furniture and household objects according to ProcTHOR's physics-grounded placement rules.

For each house, we initialize an AI2-THOR controller and retrieve reachable agent positions on the navigable floor grid. Starting near the entrance, the agent explores the house and captures exploration images at a height of 0.9\,m (a typical mobile-robot camera height) with a downward horizon tilt of $30^\circ$ and a $100^\circ$ field of view, at $512 \times 512$ resolution. The capture spacing is set adaptively from the number of reachable positions so that each house yields up to 30 RGB frames. Ground-truth observer-view images are rendered from a separate camera placed at 1.55\,m height, approximating average human eye level, with a $100^\circ$ field of view, at $512 \times 512$ resolution.

\subsection{Query Generation}

Each natural language query centers on a target object that an observer claims to have last seen at a location relative to an anchor object, while engaged with an observer object that represents their activity context. The spatial relation between the target and the anchor is drawn from four categories: right, left, front, and up.
For each validated \texttt{(observer, anchor)} pair, we generate one sentence for each of the three conditions, which vary in how much grounding information is conveyed:

\begin{itemize}[leftmargin=1em,itemsep=0.15em,topsep=0.2em]
\item \textbf{A: Inferred.}
\textit{``While replacing the bulb, I spotted the book to the left of the bed.''}. %This requires contextual and commonsense reasoning. 

\item \textbf{B: Stated.}
\textit{``While replacing the light bulb on the floor lamp, I spotted the book to the left of the bed.''}

\item \textbf{C: Given.}
\textit{``The book was to the left of the bed.''}.
\end{itemize}

Object categories are sampled from the ProcTHOR \texttt{asset types}, but sentence generation allows natural lexical variations (e.g., ``couch'' or ``sofa'' for \texttt{Sofa}) as long as the referenced furniture type remains unambiguous. The detailed object selection policies are provided in
Appendix~\ref{sec:appendix_observer_anchor_policy} and
Appendix~\ref{sec:appendix_target_policy}.

\subsection{Ground-Truth Construction}

For each (observer, anchor) pair, we render an observer-view image (Section~\ref{sec:env}) with the camera oriented toward the anchor center. A VLM is used only to judge, for each of the four spatial relations (\textsc{Right}, \textsc{Left}, \textsc{Front}, \textsc{Up}), whether target placement is physically plausible and, if so, its supporting surface. Target objects are treated as $0.4\,\text{m}$ cubes. For each valid direction, we select a target object from a surface-type lookup table and deterministically compute its ground-truth 3D coordinate; detailed placement rules are given in Appendix~\ref{app:gt}. The resulting tuples undergo two filters. We discard directions judged physically implausible by the VLM and geometrically validate each computed coordinate against the ProcTHOR floor plan, excluding coordinates that penetrate a wall or fall outside the scene boundary. The remaining ground-truth coordinates are projected into image space using a standard pinhole camera model.

\subsection{Dataset Statistics}
The benchmark comprises 47 houses and 288 observer-centric scenarios constructed from validated (observer, anchor) pairs. Each scenario contains up to four spatial directions, yielding 813 evaluation instances per condition (2,439 total across \textit{Inferred}, \textit{Stated}, and \textit{Given}).
On average, each pair yields 2.8 valid placement directions.
Among valid directions, \textsc{Front} is the most common (97\% of pairs),
followed by \textsc{Up} (66\%), \textsc{Left} (62\%), and \textsc{Right} (58\%).

In the observer-viewpoint image (\textit{Given}), 92.9\% of ground-truth targets fall within the frame; for \textit{Inferred} and \textit{Stated}, the in-frame rate depends on the exploration frame each model selects.
The benchmark dataset and evaluation code are described in Appendix~\ref{sec:appendix_reproducibility}.

\section{Experiments}
We conduct experiments across state-of-the-art models for the purposes of: (1) evaluating VLM performance on the contextual observer grounding task, and (2) investigating which grounding cues and reasoning strategies improve situated localization accuracy. We evaluate all three benchmark conditions: \textit{Inferred}, \textit{Stated}, and \textit{Given} and analyze the performance differences between them.

\subsection{Experimental Setup}
\begin{figure}[t]
  \includegraphics[width=\columnwidth]{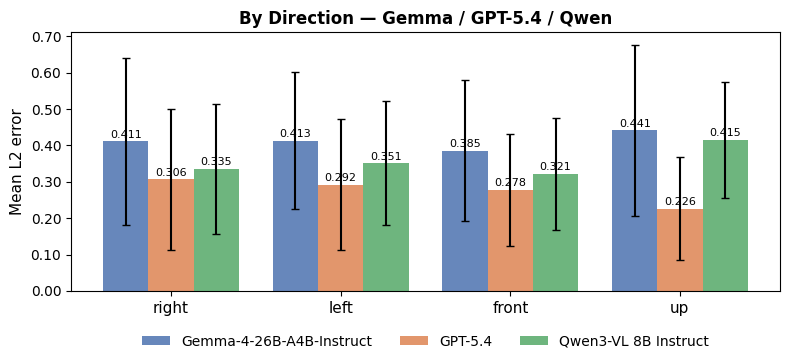}
  \caption{
Normalized L2 localization error by spatial direction on POVBench, for three representative models: Gemma-4, GPT-5.4, and Qwen3-VL-8B (Transformer).
}
  \label{fig:direction}
\end{figure}

\paragraph{Models.}
We evaluate ten VLMs spanning proprietary and open-weight general-purpose VLMs as well as dedicated pointing models \cite{clark2026molmopoint, singh2025openai, yang2025qwen3, zhu2025internvl3, team2026gemma, comanici2025gemini, team2025gemini,yuan2024robopoint,grattafiori2024llama}. Response rates are reported in Table~\ref{tab:response_rates}. Qwen3-VL-8B is tested with both Transformer and vLLM implementations. RoboPoint and Llama-3.2-Vision are evaluated only under \textit{Given}, as they did not produce valid responses in the multi-image settings. Models are accessed through their respective APIs or run locally on four NVIDIA RTX A6000 GPUs.

\paragraph{Input format.}
For the \emph{Inferred} and \emph{Stated} conditions, the model receives up to 30 indexed first-person exploration images of the house at $512{\times}512$ resolution, followed by the natural-language sentence.
The prompt additionally specifies the image coordinate system, including the top-left origin, axis directions, valid coordinate range, and image resolution. For the \emph{Given} condition, the exploration images are replaced by a single observer-viewpoint image at $512{\times}512$ resolution.

\paragraph{Prompt and reasoning protocol.}
In the \emph{Inferred} and \emph{Stated} conditions, the model is asked to (1) select the image frame that best shows both the observer object and the spatial anchor and (2) predict the pixel coordinates of the target object's center in that image, even if the object is not directly visible. Outputs are restricted to a JSON object of the form \texttt{\{selected\_image, x, y\}} without requiring intermediate reasoning traces. In
the \emph{Given} condition, the model predicts coordinates directly from the observer-viewpoint image using \texttt{\{x, y\}}. Full prompts are provided in Appendix~\ref{sec:appendix_prompts}.

\paragraph{Evaluation metrics.}
The primary metric is the normalized L2 localization error:
\begin{equation}
  \ell_2 = \sqrt{\!\left(\frac{\hat{x}}{W} - x^{\ast}\right)^{\!2}
                +\left(\frac{\hat{y}}{H} - y^{\ast}\right)^{\!2}},
\end{equation}
where $(\hat{x},\hat{y})$ is the predicted pixel coordinate,
$(W,H)$ is the image resolution, and $(x^{\ast},y^{\ast}) \in [0,1]^2$
is the ground-truth position.
For the \emph{Inferred} and \emph{Stated} conditions, the ground truth is obtained by projecting the 3D placement coordinate onto the image selected by the model, rather than the observer-viewpoint image. Both predicted and ground-truth coordinates are clamped to $[0,1]$ before computing this distance. We retain all parseable predictions, including out-of-range coordinates. Only unparseable outputs are excluded and reflected in the response rate.
We report results broken down by condition (\emph{Inferred}, \emph{Stated}, \emph{Given}), spatial direction (\textsc{Right}, \textsc{Left}, \textsc{Front}, \textsc{Up}),
and model.
We also conduct qualitative failure analysis to characterize systematic error patterns across conditions.

\subsection{Main Results}
Table~\ref{tab:main_results} summarizes normalized L2 localization errors across models and conditions.

\paragraph{Condition comparison.}
Across models evaluated on both conditions, the difference between \textit{Inferred} and \textit{Stated} is at most $0.012$ in normalized L2 error, despite \textit{Stated} explicitly naming the observer object. This small gap suggests that activity-context inference is not the sole challenge. Errors are generally lower in \textit{Given}, where the observer view is directly available, but remain substantial for most models. Together, the three conditions show that accurate spatial inference remains challenging even when the perspective is given.

\paragraph{Model comparison.}
MolmoPoint-8B achieves the lowest error in \textit{Inferred} ($0.294$) and \textit{Stated} ($0.282$), whereas GPT-5.4 performs best in \textit{Given} ($0.189$). MolmoPoint uses a dedicated pointing architecture that selects image locations from visual features, rather than generating coordinates as text. RoboPoint, another specialized pointing model, achieves the second-lowest error in \textit{Given} ($0.209$) but cannot be evaluated in the multi-image settings. These results suggest that specialized pointing mechanisms improve coordinate precision compared to typical text-token generation, yet contextual observer grounding remains challenging when the observer viewpoint must be inferred or reconstructed.

\paragraph{Direction analysis.}
Figure~\ref{fig:direction} shows L2 errors by direction for Gemma-4, GPT-5.4, and Qwen3-VL-8B (Transformer). Errors across \textsc{Right}, \textsc{Left}, and \textsc{Front} are relatively consistent, suggesting that lateral and frontal reasoning pose comparable difficulty. \textsc{Up} yields the highest errors for most models, with Gemma-4 ($0.441$) and Qwen3-VL-8B (Transformer) ($0.415$) showing notably larger errors.

\begin{table}[t]
\centering
\small
\setlength{\tabcolsep}{4pt}
\renewcommand{\arraystretch}{1.12}

\renewcommand{\tabularxcolumn}[1]{m{#1}} 
\begin{tabularx}{\columnwidth}{>{\raggedright\arraybackslash}X ccc}
\hline
\textbf{Model} & \textbf{Inferred} & \textbf{Stated} & \textbf{Given} \\
\hline
MolmoPoint-8B                  & \textbf{0.294} & \textbf{0.282} & 0.237 \\
GPT-5.4                       & 0.315 & 0.311 & \textbf{0.189} \\
Qwen3-VL-8B (Transformer)     & 0.372 & 0.369 & 0.296 \\
Qwen3-VL-8B (vLLM)            & 0.359 & 0.356 & 0.291 \\
Qwen3-VL-32B (vLLM)           & 0.449 & 0.440 & 0.488 \\
InternVL3-38B                 & 0.372 & 0.364 & 0.232 \\
Gemma-4                        & 0.463 & 0.464 & 0.367 \\
Gemini-2.5-Flash               & 0.489 & 0.487 & 0.434 \\
Gemini-Robotics-ER             & 0.535 & 0.538 & 0.507 \\
RoboPoint                      & --    & --    & 0.209 \\
Llama-3.2-Vision               & --    & --    & 0.433 \\
\hline
\end{tabularx}

\caption{
Main benchmark results on POVBench. Mean normalized L2 localization errors
(lower is better), averaged over three runs and spatial directions, computed
over answered instances. Errors are generally lowest for \textit{Given} and highest for \textit{Inferred}; the small \textit{Inferred}–\textit{Stated} gap indicates that explicitly naming the observer offers limited benefit. RoboPoint and Llama-3.2-Vision are evaluated only under the \textit{Given} condition. The best value in each condition is in bold.
}

\label{tab:main_results}
\end{table}

\subsection{Failure Analysis}

\paragraph{Contextual scene inconsistency.}
Contextual failures arise in both scene selection and anchor identification. For instance, when an instruction mentions replacing a light bulb beside a desk, models may select a desk image without any visible lamp. When the intended anchor is a refrigerator, models may mistake a nearby rectangular door frame for it rather than select the frame containing the obliquely viewed refrigerator, despite a visible toilet beyond the doorway. These errors suggest that current VLMs do not consistently integrate activity cues, visual perception, and contextual reasoning.

\paragraph{Ambiguity in vertical spatial relations.}
Up-direction predictions exhibit larger localization errors partly due
to geometric ambiguity in the evaluation setup. Many receptacle objects
(e.g., chairs with backrests) have irregular upper surfaces, making the
reference position sensitive to bounding-box definitions. In addition,
objects placed ``on'' a receptacle may vary substantially in horizontal
position while still satisfying the spatial relation. Since our
evaluation uses the anchor-object center as the reference point, such
valid positional variation can amplify measured error despite preserving
the intended relation.

\paragraph{Viewpoint estimation errors.}
Even when the correct exploration image is selected, models struggle to reconstruct the observer's facing direction and infer situated directions, leading to systematic errors in target placement estimation. This failure pattern motivates the controlled interventions evaluated in
the following section.

\paragraph{Output-format failures.}
Some models produce parseable coordinates that violate the requested $[0,511]$ range. Qwen3-VL-32B frequently follows a $[0,1000]$ convention, yielding out-of-frame $x$ coordinates for 21.0\% of \textit{Inferred} instances. Gemini-Robotics-ER also produces out-of-range coordinates across conditions. These failures demonstrate that localization error reflects not only spatial reasoning but also model-specific output behavior.

\subsection{Controlled Interventions}
To identify interventions that improve situated localization, we evaluate four visual or prompt interventions on Qwen3-VL-8B (Transformer) (Table~\ref{tab:ablation_results}): adding a visual marker for the observer position, adding markers for both the observer and anchor object, inducing chain-of-thought, and introducing spatial chain-of-thought.

\paragraph{Visual grounding cues.}
Marking only the observer's location (\textit{O-Plot}) increases error relative to the baseline, suggesting that an observer cue without the complementary anchor reference is counterproductive. Adding the anchor marker (\textit{O\&A-Plot}) restores and slightly improves performance, confirming that both reference points must be grounded together for the visual cue to be effective.

\paragraph{Structured spatial reasoning.}
Chain-of-thought reasoning improves localization, and Spatial-CoT (\textit{S-CoT}) achieves the largest improvement. Spatial-CoT explicitly guides the model through directional-frame definition, anchor localization, scene inventory, and target estimation. The improvement across both \textit{Inferred} and \textit{Stated} indicates that explicit breakdowns of observer-relative reasoning can improve localization. \textit{CoT} and \textit{S-CoT} answer fewer instances, occasionally failing to emit a parseable answer after long reasoning.

\begin{table}[t]
\centering
\small
\setlength{\tabcolsep}{4pt}
\renewcommand{\arraystretch}{1.12}

\renewcommand{\tabularxcolumn}[1]{m{#1}}
\begin{tabularx}{\columnwidth}{>{\raggedright\arraybackslash}X ccc}
\hline
\textbf{Intervention} & \textbf{Inferred} & \textbf{Stated} & \textbf{Resp.} \\
\hline
Basic      & 0.372 & 0.369 & 100\% \\
O-Plot     & 0.398 & 0.401 & 100\% \\
O\&A-Plot  & 0.342 & 0.343 & 100\% \\
CoT        & 0.322 & 0.333 & 79\%  \\
S-CoT      & \textbf{0.294} & \textbf{0.302} & 85\%  \\
\hline
\end{tabularx}
\caption{
Ablation results on POVBench (Qwen3-VL-8B (Transformer)). Mean normalized L2
errors, averaged across spatial directions and computed over answered
instances under the same protocol as Table~\ref{tab:main_results}. Resp.\ is the response rate.
}
\label{tab:ablation_results}
\end{table}

\subsection{Real-World Scene Evaluation}
\label{sec:realworld_eval}

\begin{figure}[t]
  \includegraphics[width=\columnwidth]{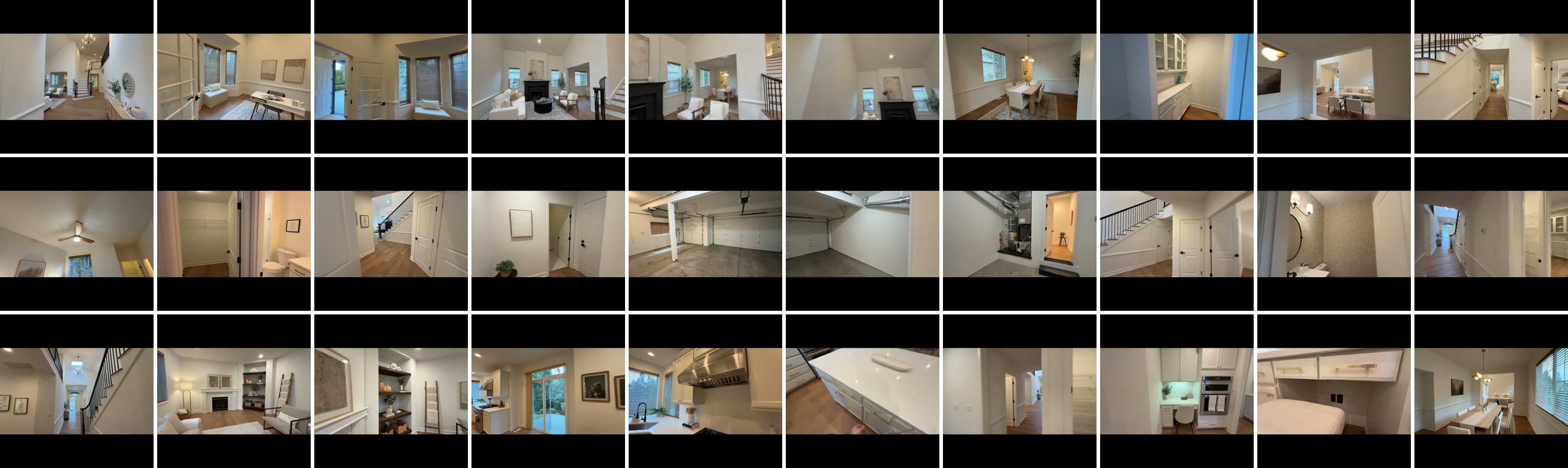}
  \caption{Thirty exploration frames extracted from the first house-tour video~\cite{noori2025housetour}.
  These images serve as the visual context for the real-world evaluation.}
  \label{fig:real_30exp}
\end{figure}

\begin{figure}[t]
  \includegraphics[width=\columnwidth]{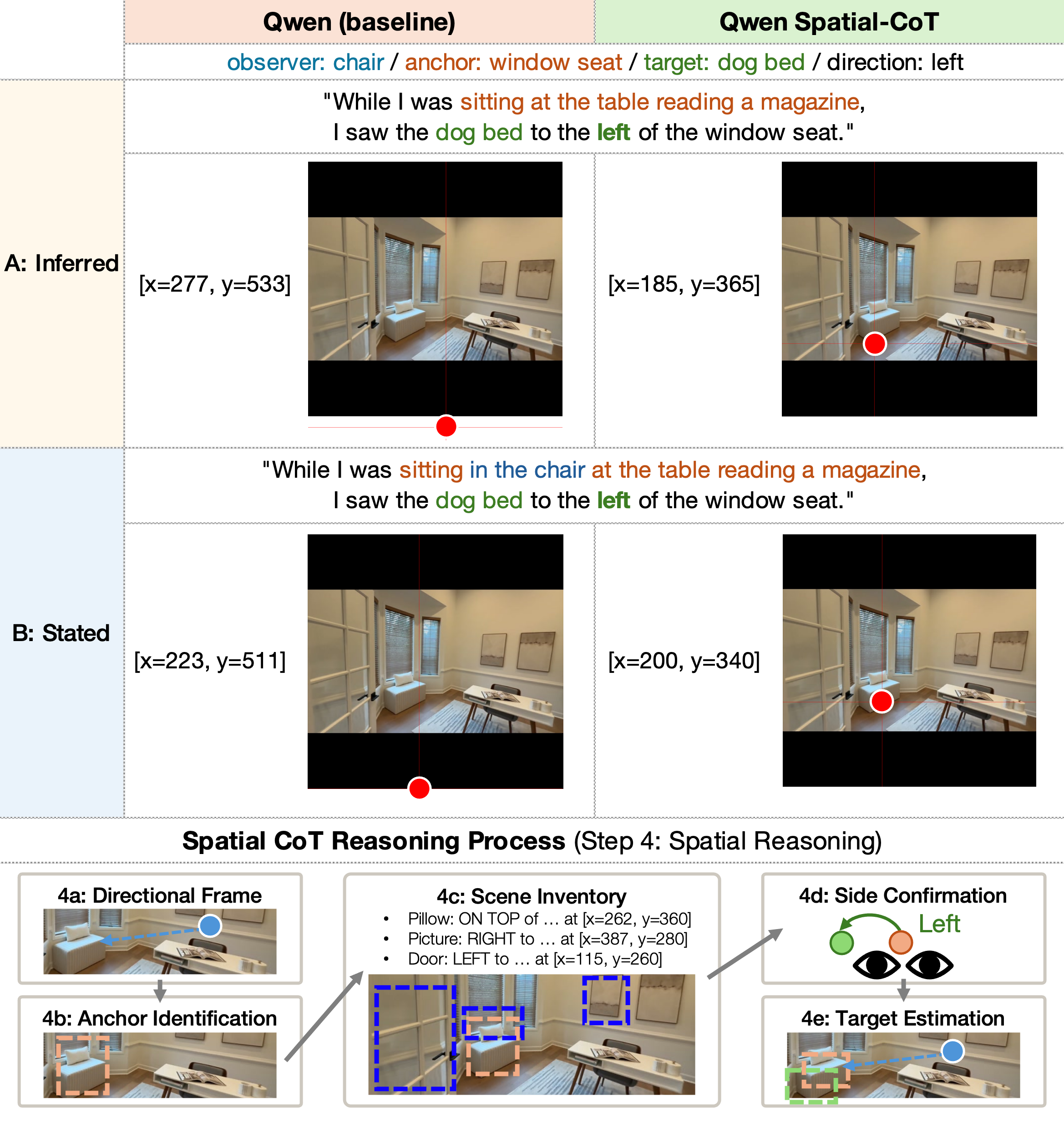}
  \caption{Qualitative comparison between standard and Spatial-CoT (\textit{S-CoT}) prompting on real-world observer-grounding scenarios. \textit{S-CoT} more reliably resolves situated spatial relations by explicitly grounding the observer viewpoint and nearby scene layout.}
  \label{fig:real_result}
\end{figure}

To test whether the reasoning challenges identified in POVBench persist in real-world settings, we conduct a small-scale validation on real house images.

\paragraph{Data.}
We use two publicly available CC BY-licensed house-tour videos~\cite{noori2025housetour,dowling2026housetour} and extract 30 exploration frames from each. For the first house, we sample frames at equal intervals ($\approx 11.5$\,s apart) throughout an approximately six-minute first-floor walkthrough, from the house entrance to the completion of a room-to-room circuit. For the second house, we uniformly sample 60 candidate frames from several minutes of indoor walkthrough footage and select a set of 30 first-floor interior frames that is as temporally contiguous as possible, discarding frames containing the presenter or outdoor areas. All frames are resized to $512 \times 512$ using letterbox padding and provide the visual context for the \textit{Inferred} and \textit{Stated} conditions. Figure~\ref{fig:real_30exp} shows the frames extracted from the first house.

We manually construct commands grounded in the real environment following the same three conditions used in POVBench. For each observer-centric scenario, we additionally select the frame judged to be closest to the observer's original viewpoint and use it as the observer-view image. In total, the evaluation contains ten observer-centric scenarios (five per house) and 102 evaluation instances spanning right, left, front, and up relations across the three conditions. Example commands are shown in Table~\ref{tab:realworld_eval_pairs}.

\begin{table}[t]
\centering
\scriptsize
\setlength{\tabcolsep}{2pt}
\renewcommand{\arraystretch}{1.15}

\begin{tabular}{
>{\centering\arraybackslash}m{0.35cm}
>{\centering\arraybackslash}m{1.25cm}
>{\raggedright\arraybackslash}m{4.0cm}
>{\centering\arraybackslash}m{0.28cm}
>{\centering\arraybackslash}m{0.28cm}
>{\centering\arraybackslash}m{0.28cm}
>{\centering\arraybackslash}m{0.28cm}
}
\hline
\textbf{Idx.} &
\textbf{View} &
\textbf{Command Examples} &
\textbf{L} &
\textbf{R} &
\textbf{F} &
\textbf{U} \\
\hline

1 &
\includegraphics[width=1.15cm]{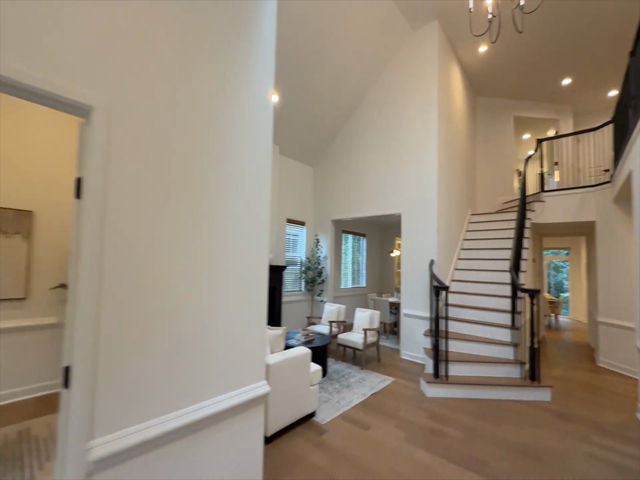}
&
\textbf{A:} While I was carefully wiping down the entrance so I wouldn't knock over the plant sitting on the books, I saw the tissuebox on the right side of the table in front of the fireplace.

\textbf{B:} While I was carefully wiping down the mirror at the entrance so I wouldn't knock over the plant sitting on the books, I saw the tissuebox on the right side of the table in front of the fireplace.

\textbf{C:} The tissuebox was on the right side of the table in front of the fireplace.
&
\checkmark & \checkmark & \checkmark & \checkmark \\
\hline

2 &
\includegraphics[width=1.15cm]{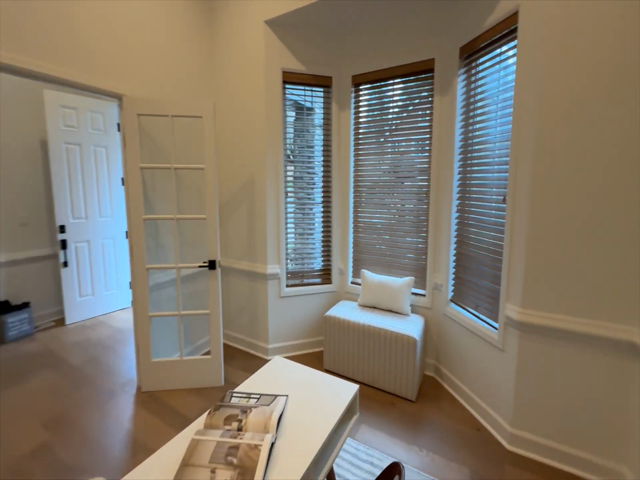}
&
\textbf{A:} While I was sitting at the table reading a magazine, I saw the dog bed to the right of the window seat.

\textbf{B:} While I was sitting in the chair at the table reading a magazine, I saw the dog bed to the right of the window seat.

\textbf{C:} The dog bed was to the right of the window seat.
&
\checkmark & \checkmark & \checkmark & \checkmark \\
\hline

3 &
\includegraphics[width=1.15cm]{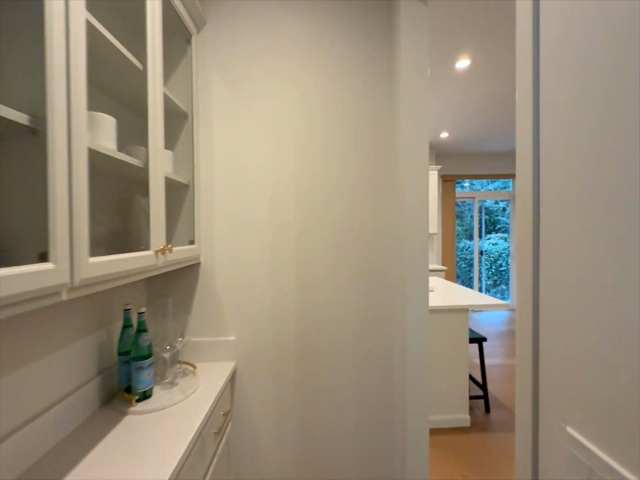}
&
\textbf{A:} While I was reaching up to put glasses away above the long white counter, I spotted the mug on the right side of the counter.

\textbf{B:} While I was reaching up to put glasses away in the cabinet above the long white counter, I spotted the mug on the right side of the counter.

\textbf{C:} The mug was on the right side of the counter.
&
\checkmark & \checkmark & \checkmark & \checkmark \\
\hline

4 &
\includegraphics[width=1.15cm]{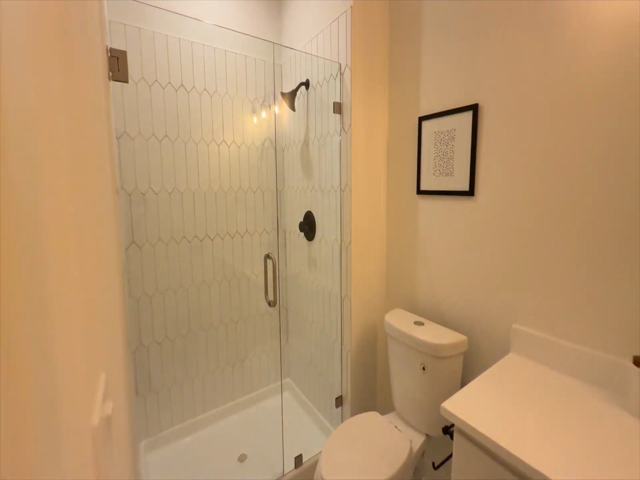}
&
\textbf{A:} While I was washing my hands in the bathroom with a painting on the wall, I saw a roll of toilet paper sitting on top of the toilet tank.

\textbf{B:} While I was washing my hands at the sink in the bathroom with a painting on the wall, I saw a roll of toilet paper sitting on top of the toilet tank.

\textbf{C:} A roll of toilet paper was sitting on top of the toilet tank.
&
 & & & \checkmark \\
\hline

5 &
\includegraphics[width=1.15cm]{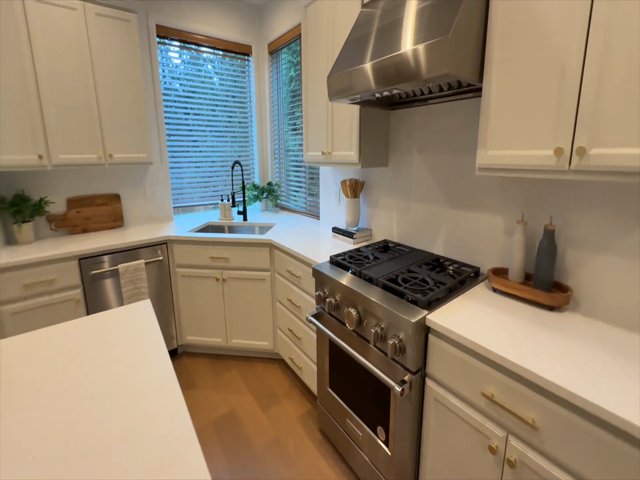}
&
\textbf{A:} While I was washing dishes, I saw a jar of jam on the left side of the stove.

\textbf{B:} While I was washing dishes at the sink, I saw a jar of jam on the left side of the stove.

\textbf{C:} A jar of jam was on the left side of the stove.
&
\checkmark & & & \\
\hline
\end{tabular}

\caption{
Real-world evaluation examples constructed from the house-tour videos~\cite{noori2025housetour,dowling2026housetour}.
Each row corresponds to one set of observer-anchor-target objects. The observer-view image becomes the reference image for \textit{C: Given} commands.
Checkmarks indicate which spatial-direction conditions were evaluated.
}
\label{tab:realworld_eval_pairs}
\end{table}

\begin{table}[!tb]
\centering
\small
\setlength{\tabcolsep}{4pt}
\renewcommand{\arraystretch}{1.12}

\renewcommand{\tabularxcolumn}[1]{m{#1}}
\begin{tabularx}{\columnwidth}{>{\raggedright\arraybackslash}X ccc}
\hline
\textbf{Model} & \textbf{Inferred} & \textbf{Stated} & \textbf{Given} \\
\hline
GPT-5.4                    & 0.29 & 0.21 & 0.41 \\
Qwen3-VL-8B (Transformer)  & 0.03 & 0.09 & 0.24 \\
Gemma-4                    & \textbf{0.41} & \textbf{0.38} & \textbf{0.47} \\
Gemini-2.5-Flash           & 0.09 & 0.09 & 0.21 \\
Gemini-Robotics-ER         & 0.29 & 0.24 & 0.24 \\
\hline
Qwen3-VL-8B + S-CoT        & 0.15 & 0.15 & --   \\
\hline
\end{tabularx}

\caption{
Manually evaluated success rates (0--1, higher is better) for real-world scene
evaluation, pooled over two house-tour videos.
Instances without a parseable prediction are counted as failures. The last row applies the \textit{S-CoT} prompt to Qwen3-VL-8B (Transformer).
}
\label{tab:realworld_results}
\end{table}

\paragraph{Models and conditions.}
We evaluate five models selected from the simulation experiments, using one run per instance. We additionally evaluate Qwen3-VL-8B (Transformer) with the best-performing \textit{S-CoT} prompt to assess whether its benefit transfers to real-world scenes.

\paragraph{Results.}
Table~\ref{tab:realworld_results} reports manually evaluated target-localization success rates across models and sentence conditions. Under the basic prompt, performance remains at or below 50\% in every condition, suggesting that contextual observer grounding remains challenging beyond the simulated environments.

For Qwen3-VL-8B (Transformer), \textit{S-CoT} improves accuracy over the basic prompt in the \textit{Inferred} and \textit{Stated} conditions. Although this small-scale evaluation is illustrative, it suggests that explicit breakdowns of observer-relative reasoning can also help in real-world scenes. Qualitative examples are shown in Figure~\ref{fig:real_result}.

\section{Conclusion and Future Work}
We introduced POVBench, a benchmark for evaluating contextual observer grounding, the ability to reason about situated spatial relations from contextual and remembered environmental information. Across ten VLMs, performance remains limited across all grounding conditions. The modest difference between \textit{Inferred} and \textit{Stated} indicates that explicitly identifying the observer provides only limited benefit. Moreover, the gains from structured observer-relative reasoning suggest that it may be a potentially key capability for embodied AI. Future work includes integrating such reasoning into embodied models and extending evaluation to diverse real-world settings.

\FloatBarrier
\section*{Limitations}
POVBench has several limitations. First, although we include a small-scale real-world evaluation with manual assessment, the benchmark is primarily constructed in ProcTHOR environments and may not capture the full complexity of real-world embodied perception and interaction.

Second, the benchmark focuses on situated spatial reasoning in indoor household environments and evaluates only four directional relations and a limited range of activity contexts. It also frames the task solely as image-space
localization; relation classification, multi-hop inference, and navigation are left to future work.

Third, the ground-truth target placements are generated semi-automatically, using a VLM to judge feasible placement and supporting surface followed by geometric filtering against the floor plan. As a result, the ground truth may contain residual noise. The point-based L2 metric also uses a single canonical target location and may penalize alternative placements that satisfy the intended spatial relation.

Fourth, we do not decompose errors into component reasoning steps: requiring an
explicit intermediate observer prediction altered the subsequent target
prediction, and we do not establish whether the Spatial-CoT gains reflect
improved observer-relative reasoning or prompt-format effects. The traces
produced by our CoT prompts make such analyses possible, and we leave a
systematic comparison across models to future work.

Fifth, our evaluation focuses on passive inference from exploration observations and does not study interactive behaviors such as dialogue-based clarification or embodied exploration policies. Future work should extend evaluation to more diverse real-world settings, richer contextual grounding scenarios, and interactive embodied agents.

% This document does not cover the content requirements for ACL or any
% other specific venue.  Check the author instructions for
% information on
% maximum page lengths, the required ``Limitations'' section,
% and so on.

% \section*{Acknowledgments}
% This work was supported by ...
% This document has been adapted
% by Steven Bethard, Ryan Cotterell and Rui Yan
% from the instructions for earlier ACL and NAACL proceedings, including those for
% ACL 2019 by Douwe Kiela and Ivan Vuli\'{c},
% NAACL 2019 by Stephanie Lukin and Alla Roskovskaya,
% ACL 2018 by Shay Cohen, Kevin Gimpel, and Wei Lu,
% NAACL 2018 by Margaret Mitchell and Stephanie Lukin,
% Bib\TeX{} suggestions for (NA)ACL 2017/2018 from Jason Eisner,
% ACL 2017 by Dan Gildea and Min-Yen Kan,
% NAACL 2017 by Margaret Mitchell,
% ACL 2012 by Maggie Li and Michael White,
% ACL 2010 by Jing-Shin Chang and Philipp Koehn,
% ACL 2008 by Johanna D. Moore, Simone Teufel, James Allan, and Sadaoki Furui,
% ACL 2005 by Hwee Tou Ng and Kemal Oflazer,
% ACL 2002 by Eugene Charniak and Dekang Lin,
% and earlier ACL and EACL formats written by several people, including
% John Chen, Henry S. Thompson and Donald Walker.
% Additional elements were taken from the formatting instructions of the \emph{International Joint Conference on Artificial Intelligence} and the \emph{Conference on Computer Vision and Pattern Recognition}.

\bibliography{custom}

\appendix
\section*{Appendix}
\section{Reproducibility}
\label{sec:appendix_reproducibility}
Our project page is available at \url{https://mimo-owl.github.io/POVBench/}.

\section{POVBench Implementation Details}
\FloatBarrier

\subsection{Room Number Distribution}
\label{sec:appendix_stats}

\begin{figure}[t]
    \centering
    \includegraphics[width=0.8\linewidth]{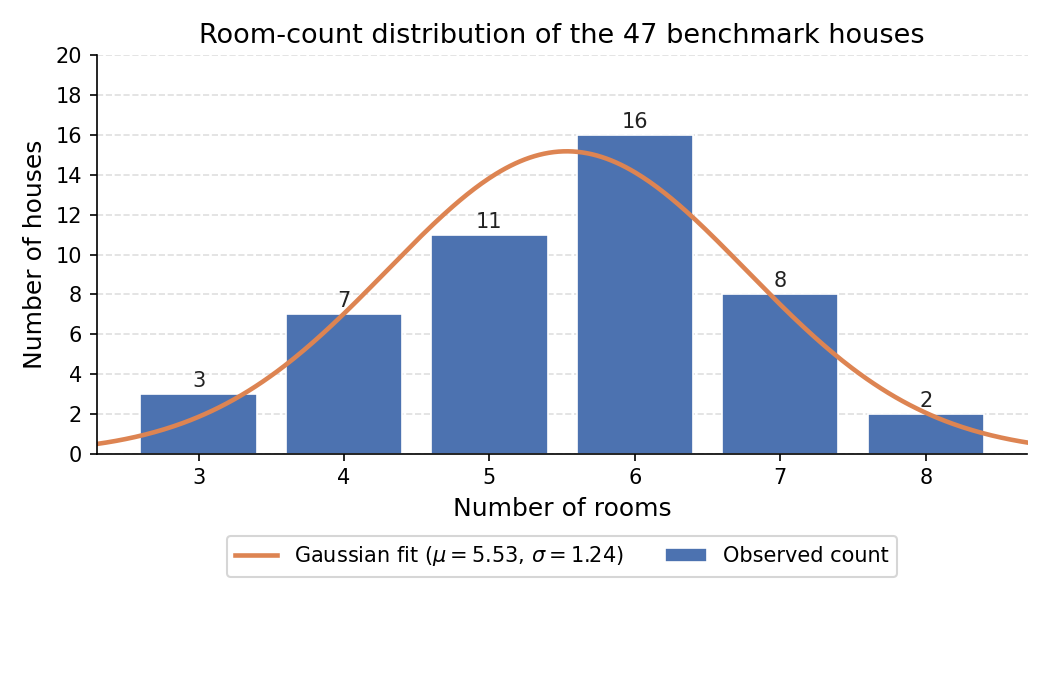}
    \caption{
    Distribution of room counts across the 47 benchmark houses. The red curve shows a fitted Gaussian distribution illustrating variability in environment complexity.
    }
    \label{fig:room_dist}
\end{figure}

Figure~\ref{fig:room_dist} shows the distribution of room counts across the benchmark houses.

\subsection{Observer / Anchor Object Categories}
\label{sec:appendix_observer_anchor_policy}
We provide the 31 candidate object categories used for observer and anchor objects in POVBench in Table~\ref{tab:observer_anchor_objects}.

\begin{table}[t]
\centering
\small

\begin{tabular}{p{0.48\linewidth} p{0.48\linewidth}}

$\bullet$ ArmChair        & $\bullet$ GarbageCan \\

$\bullet$ Bed             & $\bullet$ HandTowelHolder \\

$\bullet$ Chair           & $\bullet$ HousePlant \\

$\bullet$ ClothesDryer    & $\bullet$ Microwave \\

$\bullet$ CoffeeMachine   & $\bullet$ Safe \\

$\bullet$ CoffeeTable     & $\bullet$ ShelvingUnit \\

$\bullet$ CounterTop      & $\bullet$ SideTable \\

$\bullet$ Desk            & $\bullet$ Sink \\

$\bullet$ DeskLamp        & $\bullet$ Sofa \\

$\bullet$ Desktop         & $\bullet$ Stool \\

$\bullet$ DiningTable     & $\bullet$ Television \\

$\bullet$ Doorway         & $\bullet$ Toaster \\

$\bullet$ Dresser         & $\bullet$ Toilet \\

$\bullet$ Faucet          & $\bullet$ TowelHolder \\

$\bullet$ FloorLamp       & $\bullet$ WashingMachine \\

$\bullet$ Fridge          & \\

\end{tabular}

\caption{
31 object categories used as observer and anchor objects in the benchmark.
}

\label{tab:observer_anchor_objects}

\end{table}

\subsection{Target Object Selection Policy}
\label{sec:appendix_target_policy}

Target objects are selected according to the physical support surface associated with the sampled spatial relation. If the target object location lies on the floor, the object is sampled from a predefined floor-compatible object pool. Otherwise, it is sampled from a predefined receptacle-specific object pool associated with the supporting surface (Table~\ref{tab:surface_targets}).

For example:

\begin{itemize}[leftmargin=1em,itemsep=0.15em,topsep=0.2em]

\item If the anchor is a \texttt{Bed} and the sampled relation places the target on the floor beside the bed, the target is sampled from the floor-compatible object pool.

\item If the anchor is a \texttt{Bed} and the sampled relation places the target on top of the bed, the target is sampled from the \texttt{Bed}-specific object pool.

\item If the anchor is a \texttt{Microwave} and the sampled relation places the target on the surrounding countertop surface, the target is sampled from the \texttt{CounterTop} object pool.

\end{itemize}

% \paragraph{Receptacle-specific target objects.}

\begin{table}[t]
\centering
% \scriptsize
\small
\renewcommand{\arraystretch}{1.1}

\begin{tabularx}{\columnwidth}{p{1.8cm} X}
\hline
Surface & Candidate Target Objects \\
\hline

Floor &
Cloth, Boots, BasketBall, BaseballBat, Dumbbell, DogBed \\

ArmChair &
Book, TeddyBear, Cloth, TissueBox, Newspaper, Mug, Cup, CellPhone, RemoteControl, Watch \\

Bed &
Book, Pillow, TeddyBear, Cloth, TissueBox, Newspaper, CellPhone, RemoteControl, Watch \\

Chair &
Cloth, Book, Pen, Pencil, CellPhone, KeyChain, Watch \\

CoffeeTable &
Book, TissueBox, Newspaper, Mug, Cup, Bottle, RemoteControl, CellPhone, Watch, KeyChain \\

CounterTop &
Mug, Cup, Bottle, Plate, Bowl, Fork, Spoon, Knife, TissueBox \\

Desk &
Book, TissueBox, Newspaper, Mug, Cup, Pen, Pencil, CellPhone, Watch, KeyChain \\

DiningTable &
Plate, Bowl, Fork, Spoon, Knife, Mug, Cup, Bottle, TissueBox, Cloth, KeyChain, Watch \\

Dresser &
Cloth, Watch, KeyChain, CellPhone, Book, TissueBox \\

Microwave &
Plate, Bowl, Mug, Cup, Fork, Spoon, Newspaper, Watch, KeyChain \\

ShelvingUnit &
Book, TeddyBear, TissueBox, Newspaper, Cloth \\

SideTable &
Book, TissueBox, Mug, Cup, CellPhone, Watch, RemoteControl, KeyChain \\

Sofa &
Book, Pillow, TeddyBear, Cloth, TissueBox, Newspaper, Mug, Cup, CellPhone, RemoteControl, Watch \\

Television &
RemoteControl, CellPhone, TeddyBear, Cloth, TissueBox \\

Toilet &
TissueBox, Book, Newspaper, CellPhone \\

\hline
\end{tabularx}

\caption{
Examples of surface-specific target object pools.
Target candidates are selected according to the physical support surface associated with the sampled spatial relation.
}
\label{tab:surface_targets}

\end{table}

\subsection{Ground-Truth Placement Rules}
\label{app:gt}

All spatial relations are defined from the observer's perspective when
facing the anchor object.
Let $\mathbf{c} = (c_x, c_z)$ denote the horizontal center of the
anchor object's axis-aligned bounding box (AABB),
$y_{\mathrm{top}}$ its maximum $y$-coordinate,
and $\hat{\mathbf{d}} \in \mathbb{R}^2$ the unit direction vector in
the $xz$-plane corresponding to the spatial relation
(\textsc{Front}: anchor $\to$ observer;
 \textsc{Left}/\textsc{Right}: $\pm 90^{\circ}$ rotation of \textsc{Front}).
Let $\mathbf{b}(\hat{\mathbf{d}})$ denote the point where the ray from
$\mathbf{c}$ in direction $\hat{\mathbf{d}}$ intersects the anchor AABB
boundary.

\paragraph{Surface height.}
The VLM surface judgment yields a phrase of the form
\textit{``on the \textless surface\textgreater''}.
The surface height $y_{\mathrm{surf}}$ is set to $0$ for \textit{floor},
to $y_{\mathrm{top}}$ of the anchor for the anchor object itself,
and to the $y_{\mathrm{top}}$ of the named object otherwise.
For flat-topped anchor types (e.g., \textit{Bed}, \textit{DiningTable},
\textit{CounterTop}), \textsc{Left}/\textsc{Right} placement on the
anchor surface is also permitted, in which case $(x,z)$ is offset
laterally from $\mathbf{c}$ rather than from the AABB boundary. The placement rules for each spatial relation are summarized in Table~\ref{tab:placement_rules}. The $0.2$\,m offset reflects the assumed $0.4$\,m cube
extent of the target object (half-width placed at the surface edge or center).

\begin{table}[t]
\centering
\small
\renewcommand{\arraystretch}{1.2}

\resizebox{\linewidth}{!}{
\begin{tabular}{llll}
\hline
\textbf{Relation} & \textbf{Anchor} & \textbf{$(x,z)$} & \textbf{$y$} \\
\hline

\textsc{Up} & any
& $\mathbf{c}$
& $y_{\mathrm{top}} + 0.2$ \\

\textsc{Front} & any
& $\mathbf{b}(\hat{\mathbf{d}}) + 0.2\,\hat{\mathbf{d}}$
& $y_{\mathrm{surf}} + 0.2$ \\

\textsc{Left/Right} & $\notin \mathcal{G}$
& $\mathbf{b}(\hat{\mathbf{d}}) + 0.2\,\hat{\mathbf{d}}$
& $y_{\mathrm{surf}} + 0.2$ \\

\textsc{Left/Right} & $\in \mathcal{G}$, on anchor
& $\mathbf{c} + 0.2\,\hat{\mathbf{d}}$
& $y_{\mathrm{top}} + 0.2$ \\

\textsc{Left/Right} & $\in \mathcal{G}$, not on anchor
& $\mathbf{b}(\hat{\mathbf{d}}) + 0.2\,\hat{\mathbf{d}}$
& $y_{\mathrm{surf}} + 0.2$ \\

\hline
\end{tabular}
}

\caption{Target placement rules for each spatial relation.}
\label{tab:placement_rules}

\end{table}

\subsection{Evaluation Prompts}
\label{sec:appendix_prompts}
We provide the full prompts used in our experiments. Placeholders (\textit{N} images, the task \textit{sentence}, and image size $W\times H$) are filled at runtime. The \textbf{O-Plot} and \textbf{O\&A-Plot} interventions use the same prompt text as the basic \textit{Inferred}/\textit{Stated} prompt below, differing only in that visual markers are overlaid on the images.

\paragraph{Basic prompt (Inferred / Stated).}
\begin{tcolorbox}[breakable, colback=gray!8, colframe=gray!60, boxrule=0.5pt, arc=2mm, left=2mm, right=2mm, top=1mm, bottom=1mm]
\small
You are the intelligent brain system of a home-assistance robot. You are given first-person perspective images captured during the robot's exploration of a home.\\[0.3em]
The \textit{N} images above (labeled Image 0 through Image \textit{N}$-1$) were all taken during exploration of the same single-story house. Each image is preceded by its label (``Image 0:'', ``Image 1:'', etc.).\\[0.3em]
Sentence: \textit{``\textless sentence\textgreater''}\\[0.3em]
The resident has asked the robot to retrieve an object and is describing where they last saw it. Your mission is to predict, as accurately as possible, where that object is located, based on the resident's description in the sentence above.\\[0.5em]
\textbf{Task:}
\begin{enumerate}[leftmargin=1.4em,itemsep=0.2em,topsep=0.3em]
\item Select the image (0 to \textit{N}$-1$) that best represents the scene described in the sentence---the image where both the observer's furniture and the reference landmark are most clearly visible.
\item In the selected image, predict the center of the target object as pixel coordinates. The target object may not actually be visible in the image. Based on the sentence, predict as faithfully and accurately as possible where the center of the object \textsc{would} be located.
\end{enumerate}
\textit{Image coordinate system:} origin \texttt{[0,0]} at the top-left corner; $x$-axis points right ($x\in[0,W\!-\!1]$); $y$-axis points down ($y\in[0,H\!-\!1]$); image size $W\times H$ pixels.\\[0.3em]
IMPORTANT: When predicting coordinates, pay close attention to the origin position and axis directions, and ensure all values fall within the valid range.\\[0.3em]
Output ONLY valid JSON, no other text:\\
\texttt{\{"selected\_image": <integer 0-(N-1)>, "x": <integer 0-(W-1)>, "y": <integer 0-(H-1)>\}}
\end{tcolorbox}

\paragraph{Basic prompt (Given).}
\begin{tcolorbox}[breakable, colback=gray!8, colframe=gray!60, boxrule=0.5pt, arc=2mm, left=2mm, right=2mm, top=1mm, bottom=1mm]
\small
You are the intelligent brain system of a home-assistance robot.\\[0.3em]
Sentence: \textit{``\textless sentence\textgreater''}\\[0.3em]
The resident has asked the robot to retrieve an object and is describing where they last saw it. Your mission is to predict, as accurately as possible, where that object is located---based on the resident's description in the sentence above.\\[0.3em]
The image provided shows the target scene from the perspective of the resident at the moment they spotted the target object.\\[0.5em]
\textbf{Task:} Based on the sentence, predict the center of the target object as pixel coordinates. The target object may not actually be visible in the image. Based on the sentence, predict as faithfully and accurately as possible where the center of the object \textsc{would} be located.\\[0.3em]
\textit{Image coordinate system:} origin \texttt{[0,0]} at the top-left corner; $x$-axis points right ($x\in[0,W\!-\!1]$); $y$-axis points down ($y\in[0,H\!-\!1]$); image size $W\times H$ pixels.\\[0.3em]
IMPORTANT: When predicting coordinates, pay close attention to the origin position and axis directions, and ensure all values fall within the valid range.\\[0.3em]
Output ONLY valid JSON, no other text:\\
\texttt{\{"x": <integer 0-(W-1)>, "y": <integer 0-(H-1)>\}}
\end{tcolorbox}

\begin{table*}[t]
\centering
\small
\setlength{\tabcolsep}{6pt}
\renewcommand{\arraystretch}{1.08}
\begin{tabularx}{\textwidth}{
>{\raggedright\arraybackslash}p{0.25\textwidth}
>{\raggedright\arraybackslash}X
c}
\hline
\textbf{Paper name} & \textbf{Model identifier or endpoint}
& \textbf{Conditions} \\
\hline
MolmoPoint-8B
& \texttt{allenai/MolmoPoint-8B}
& A, B, C \\

GPT-5.4
& \texttt{gpt-5.4}
& A, B, C \\

Qwen3-VL-8B (Transformer)
& \texttt{Qwen/Qwen3-VL-8B-Instruct}
& A, B, C \\

Qwen3-VL-8B (vLLM)
& \texttt{Qwen/Qwen3-VL-8B-Instruct}
& A, B, C \\

Qwen3-VL-32B (vLLM)
& \texttt{Qwen/Qwen3-VL-32B-Instruct}
& A, B, C \\

InternVL3-38B
& \texttt{OpenGVLab/InternVL3-38B} (8-bit quantized)
& A, B, C \\

Gemma-4
& \texttt{gemma-4-26b-a4b-it}
& A, B, C \\

Gemini-2.5-Flash
& \texttt{gemini-2.5-flash}
& A, B, C \\

Gemini-Robotics-ER
& \texttt{gemini-robotics-er-1.6-preview}
& A, B, C \\

RoboPoint
& \texttt{wentao-yuan/robopoint-v1-vicuna-v1.5-13b}
& C only \\

Llama-3.2-Vision
& \texttt{meta-llama/Llama-3.2-11B-Vision-Instruct}
& C only \\
\hline
\end{tabularx}

\caption{
Model configurations used in the POVBench experiments. Qwen3-VL-8B is
evaluated using both Transformers and vLLM implementations. InternVL3-38B is
evaluated with 8-bit quantization. RoboPoint and Llama-3.2-Vision are evaluated
only under the Given condition because they do not produce valid responses in
the multi-image settings.
}
\label{tab:model_configurations}
\end{table*}

\paragraph{CoT prompt (Inferred / Stated).}
\begin{tcolorbox}[breakable, colback=gray!8, colframe=gray!60, boxrule=0.5pt, arc=2mm, left=2mm, right=2mm, top=1mm, bottom=1mm]
\small
You are the intelligent brain system of a home-assistance robot. You are given first-person perspective images captured during the robot's exploration of a home.\\[0.3em]
The \textit{N} images above (labeled Image 0 through Image \textit{N}$-1$) were all taken during exploration of the same single-story house. Each image is preceded by its label (``Image 0:'', ``Image 1:'', etc.).\\[0.3em]
Sentence: \textit{``\textless sentence\textgreater''}\\[0.3em]
The resident has asked the robot to retrieve an object and is describing where they last saw it. Your mission is to predict, as accurately as possible, where that object is located, based on the resident's description in the sentence above.\\[0.5em]
\textbf{Think step by step:}\\[0.5em]
\textbf{Step 1 --- Image selection.} Among the \textit{N} images, identify the one that best represents the scene described in the sentence---the image where both the furniture the observer was working with and the spatial landmark mentioned in the sentence are most clearly visible. Note its index.\\[0.4em]
\textbf{Step 2 --- Observer position.} In the selected image, identify the furniture the observer was working with at the time. Note roughly where it appears and what direction the observer would be facing.\\[0.4em]
\textbf{Step 3 --- Anchor landmark.} Identify the spatial landmark mentioned in the sentence in the selected image.\\[0.4em]
\textbf{Step 4 --- Spatial reasoning.} Based on the spatial relationship described in the sentence, estimate where the target object would be located in the selected image.\\[0.4em]
\textbf{Step 5 --- Conclusion.} State your selected image index and final coordinate estimate.\\[0.5em]
\textit{Image coordinate system:} origin \texttt{[0,0]} at the top-left corner; $x$-axis points right ($x\in[0,W\!-\!1]$); $y$-axis points down ($y\in[0,H\!-\!1]$); image size $W\times H$ pixels.\\[0.3em]
IMPORTANT: When predicting coordinates, pay close attention to the origin position and axis directions, and ensure all values fall within the valid range.\\[0.3em]
Write your step-by-step reasoning freely. End your response with ONLY the final JSON on the last line:\\
\texttt{\{"selected\_image": <integer 0-(N-1)>, "x": <integer 0-(W-1)>, "y": <integer 0-(H-1)>\}}
\end{tcolorbox}

\paragraph{Spatial-CoT prompt (Inferred / Stated).}
The Spatial-CoT prompt is identical to the CoT prompt above, except that \textbf{Step~4} is replaced with the following structured spatial-reasoning procedure.
\begin{tcolorbox}[breakable, colback=gray!8, colframe=gray!60, boxrule=0.5pt, arc=2mm, left=2mm, right=2mm, top=1mm, bottom=1mm]
\small
\textbf{Step 4 --- Spatial reasoning}\\[0.4em]
\textbf{4a) Directional frame.} Define the viewing direction as from the observer's furniture toward the anchor object. Interpret spatial relations from the observer's perspective while facing the anchor object:
\begin{itemize}[leftmargin=1em,itemsep=0.2em,topsep=0.3em]
\item {LEFT / RIGHT}: object lies to the left or right of the anchor object
\item {FRONT}: object is closer to the observer than the anchor object
\item {ON TOP}: object is resting on or above the anchor object
\item {OTHER}: relation does not clearly fit the above categories
\end{itemize}
\textbf{4b) Anchor coordinate.} Estimate the pixel coordinate $(x, y)$ of the anchor object.\\[0.3em]
\textbf{4c) Scene inventory.} Identify two or three nearby visible objects around the anchor object. For each object, provide: object name or description; spatial relation to the anchor object; estimated pixel coordinate $(x, y)$.\\[0.3em]
\textbf{4d) Target relation.} Determine the spatial relation of the target object relative to the anchor object.\\[0.3em]
\textbf{4e) Target coordinate estimation.} Estimate the target pixel coordinate using the anchor object coordinate, nearby object coordinates, and spacing and scale cues.
\end{tcolorbox}

\subsection{Model Configurations}
\label{app:model_details}

Table~\ref{tab:model_configurations} lists the exact model identifiers or API
endpoints used for each evaluated model, along with the conditions under which
each model is evaluated.

\subsection{Response Rates}
\label{app:response_rates}

Table~\ref{tab:response_rates} reports the response rate of each model under
every condition. A response is counted when the model returns a parseable coordinate prediction.
Requests that yield unparseable output or fail at the API level are both treated
as non-responses.

\begin{table}[t]
\centering
\scriptsize
\setlength{\tabcolsep}{3pt}
\renewcommand{\arraystretch}{1.08}
\renewcommand{\tabularxcolumn}[1]{m{#1}}
\begin{tabularx}{\columnwidth}{>{\raggedright\arraybackslash}X cccc}
\hline
\textbf{Model} & \textbf{Inferred} & \textbf{Stated} & \textbf{Given} & \textbf{Overall} \\
\hline
MolmoPoint-8B             & 97.7  & 98.5  & 98.9  & 98.4  \\
GPT-5.4                   & 97.6  & 97.6  & 97.6  & 97.6  \\
Qwen3-VL-8B (Transformer) & 100.0 & 100.0 & 100.0 & 100.0 \\
Qwen3-VL-8B (vLLM)        & 100.0 & 100.0 & 100.0 & 100.0 \\
Qwen3-VL-32B (vLLM)       & 100.0 & 100.0 & 94.6  & 98.2  \\
InternVL3-38B             & 100.0 & 100.0 & 100.0 & 100.0 \\
Gemma-4                   & 39.7  & 40.0  & 99.3  & 59.7  \\
Gemini-2.5-Flash          & 95.7  & 96.1  & 96.0  & 95.9  \\
Gemini-Robotics-ER        & 94.9  & 95.2  & 95.2  & 95.1  \\
RoboPoint                 & --    & --    & 100.0 & 100.0 \\
Llama-3.2-Vision          & --    & --    & 100.0 & 100.0 \\
\hline
\end{tabularx}

\caption{
Response rates (\%) on POVBench. Overall is the average over the conditions
each model is evaluated under. A response is counted when the model produces a
parseable coordinate prediction.
}
\label{tab:response_rates}
\end{table}

% % This is an appendix.

\end{document}